%% file: paper.tex
\documentclass[11pt]{article}
\usepackage{eamt18}\usepackage{times}
\usepackage{url}
\usepackage{latexsym}
\usepackage[small,bf]{caption}
\usepackage{booktabs} 
\usepackage[ruled]{algorithm2e}
\usepackage{algorithmic}
\usepackage{color}
\usepackage{amsmath}
\usepackage{amssymb}
\usepackage{makecell}

\newcommand{\E}{\mathbb{E}}

\title{A Reinforcement Learning Approach \\to Interactive-Predictive Neural Machine Translation}

\author{Tsz Kin Lam$^{\dagger,\ast}$ \and Julia Kreutzer$^{\ast}$ \and Stefan Riezler$^{\dagger,\ast}$ \\
  $^{\ast}$Computational Linguistics \& $^\dagger$IWR, Heidelberg University, Germany \\
  {\tt \{lam,kreutzer,riezler\}@cl.uni-heidelberg.de}\\
  }

\date{}

\begin{document}
\maketitle
\begin{abstract}
  We present an approach to interactive-predictive neural machine translation that attempts to reduce human effort from three directions: Firstly, instead of requiring humans to select, correct, or delete segments, we employ the idea of learning from human reinforcements in form of judgments on the quality of partial translations. Secondly, human effort is further reduced by using the entropy of word predictions as uncertainty criterion to trigger feedback requests. Lastly, online updates of the model parameters after every interaction allow the model to adapt quickly. We show in simulation experiments that reward signals on partial translations significantly improve character F-score and BLEU compared to feedback on full translations only, while human effort can be reduced to an average number of $5$ feedback requests for every input.
\end{abstract}

\section{Introduction}
\label{sec:intro}
\input{sec-intro.tex}

\section{Related Work}
\label{sec:related}
\input{sec-related.tex}

\section{Reinforcement Learning for Interactive-Predictive Translation}
\label{sec:interactive}
\input{sec-interactive.tex}

\section{Experiments}
\label{sec:exps}
\input{sec-exps.tex}

\section{Conclusion}
\label{sec:disc}
\input{sec-disc.tex}

\section*{Acknowledgments.} This work was supported in part by DFG Research Grant RI 2221/4-1. We would like to thank the members of the Statistical NLP Colloquium Heidelberg for fruitful discussions and ideas for improvement of our algorithm.

\bibliography{references}
\bibliographystyle{eamt18}

\clearpage

\end{document}

%% file: sec-intro.tex
Interactive-predictive machine translation aims at obtaining high-quality machine translation by involving humans in a loop of user validations of partial translations suggested by the machine translation system. This interaction protocol can easily be fit to neural machine translation (NMT) \cite{BahdanauETAL:15} by conditioning the model's word predictions on the user-validated prefix \cite{KnowlesKoehn:16,WuebkerETAL:16}. User studies conducted by Green et al. \shortcite{GreenETAL:14} for phrase-based machine translation have shown that the interactive-predictive interaction protocol leads to significant reductions in post-editing effort. Other user studies on interactive machine translation based on post-editing have shown that human effort can also be reduced by improving the online adaptation capabilities of the learning system, both for statistical phrase-based \cite{BentivogliETAL:16} or NMT systems \cite{KarimovaETAL:17}.

The goal of our work is to further reduce human effort in interactive-predictive NMT by combining the advantages of the interactive-predictive protocol with the advantages of learning from weak feedback. For the latter we rely on techniques from reinforcement learning \cite{SuttonBarto:17}, a.k.a. bandit structured prediction \cite{SokolovETALnips:16,KreutzerETAL:17,NguyenETAL:17} in the context of sequence-to-sequence learning. Our approach attacks the problem of reducing human effort from three innovative directions.
\begin{itemize}
\item Firstly, instead of requiring humans to correct or delete segments proposed by the machine translation system, we employ the reinforcement learning idea of humans providing reward signals in form of judgments on the quality of the machine translation. Human effort is reduced since each partial translation receives a human reward signal at most once, rendering it a bandit-type feedback signal, and each reward signal itself is easier to obtain than a correction of a translation.
\item In order to reduce the amount of feedback signals even further, we integrate an uncertainty criterion for word predictions to trigger requests for human feedback. Using the comparison of the current average entropy to the entropy of word predictions in the history as a measure for uncertainty, we reduce the amount of feedbacks requested from humans to an average number of $5$ requests per input.
\item In contrast to previous approaches to interactive-predictive translation, the parameters of our translation system are updated online after receiving feedback for partial translations. The update is done according to an actor-critic reinforcement learning protocol where each update pushes up the score function of the partial translation sampled by the model (called actor) proportional to a learned reward function (called critic). Furthermore, since the entropy criterion is based on the actor, it is also automatically updated. Frequent updates improve the adaptability of our system, resulting in a further reduction of human effort.
\end{itemize}

The rest of this paper is structured as follows. In Section \ref{sec:related}, we will situate our approach in the context of interactive machine translation and analyze our contribution related to reinforcement learning for sequence prediction problems. Details of our algorithm are given in Section \ref{sec:interactive}. We evaluate our approach in a simulation study where bandit feedback is computed by evaluating partial translations against references under a character F-score metric \cite{Popovic:15} without revealing the reference translation to the learning system (Section \ref{sec:exps}). We show that segment-wise reward signals improve translation quality over reinforcement learning with sparse sentence-wise rewards, measured by character F-score and corpus-based BLEU against references. Furthermore, we show that human effort, measured by the number of feedback requests, can be reduced to an average number of $5$ requests per input. These implications of our new paradigm are discussed in Section \ref{sec:disc}.

%% file: sec-related.tex
The interactive-predictive translation paradigm reaches back to early approaches for IBM-type \cite{FosterETAL:97,FosterETAL:02} and phrase-based machine translation \cite{BarrachinaETAL:08,GreenETAL:14}. Knowles and Koehn \shortcite{KnowlesKoehn:16} and Wuebker et al. \shortcite{WuebkerETAL:16} presented \emph{neural interactive translation prediction} --- a translation scenario where translators interact with an NMT system by accepting or correcting subsequent target tokens suggested by the NMT system in an auto-complete style. NMT is naturally suited for this incremental production of outputs, since it models the probability of target tokens given a history of target tokens sequentially from left to right. In standard supervised training with teacher forcing, this history comes from the ground truth, 
while in interactive-predictive translation it is provided by the prefix accepted or entered by the user. Both approaches use references to simulate an interaction with a translator and compare their approach to phrase-based prefix-search. They find that NMT is more accurate in word and letter prediction and recovers better from failures. Similar to their work, we will experiment in a simulated environment with references mimicking the translator. However, we do not use the reference directly for teacher forcing, but only to derive weak feedback from it. Furthermore, our approach employs techniques to reduce the number of interactions, and to update the model more frequently than after each sentence. 

Our work is also closely related to approaches for \emph{interactive pre-post-editing} \cite{MarieMax:15,DomingoETAL:18}. The core idea is to ask the translator to mark good segments and use these for a more informed re-decoding. Both studies could show a reduction in human effort for post-editing in simulation experiments. We share the goal of using human feedback more effectively by targeting it towards essential translation segments, however, our approach does adhere to the left-to-right navigation through translation hypotheses. In difference to these approaches, we try to reduce human effort even further by minimizing the number of feedback requests and by frequent model updates.

Reinforcing/penalizing a targeted set of actions can also be found in recent approaches to \emph{reinforcement learning from human feedback}. For example, Judah et al. \shortcite{JudahETAL:10} presented a scenario where users interactively label freely chosen good and bad parts of a policy's trajectory. The policy is directly trained with this reinforcement signal to play a real-time strategy game. Simulations of NMT systems interacting with human feedback have been presented firstly by Kreutzer et al. \shortcite{KreutzerETAL:17}, Nguyen \shortcite{NguyenETAL:17}, or Bahdanau et al. \shortcite{BahdanauETAL:17} who apply different policy gradient algorithms, William's REINFORCE \cite{Williams:92} or actor-critic methods \cite{KondaTsitsiklis:00,SuttonETAL:00,MnihETAL:16}, respectively. While Bahdanau et al.'s \shortcite{BahdanauETAL:17} approach operates in a fully supervised learning scenario, where rewards are simulated in comparison to references with smoothed and length-rescaled BLEU, Kreutzer et al. \shortcite{KreutzerETAL:17} and Nguyen et al. \shortcite{NguyenETAL:17} limit the setup to sentence-level bandit feedback, i.e. only one feedback is obtained for one completed translation per input. In this paper, we use actor-critic update strategies, but we receive simulated bandit feedback on the sub-sentence level.

We adopt techniques from \emph{active learning} to reduce the number of feedbacks requested from a user. 
Gonz{\'a}lez-Rubio et al. \shortcite{Gonzalez-RubioETAL:11,Gonzalez-RubioETAL:12} apply active learning for interactive machine translation, where a user interactively finishes the translation of an SMT system. The active learning component decides which  sentences to sample for translation (i.e. receive full supervision for) and the SMT system is updated online \cite{Ortiz-MartinezETAL:10}. In our algorithm the active learning component decides which prefixes to be rated (i.e. receive weak feedback for) based on their average entropy. Entropy is a popular measure for uncertainty in active learning: the rationale is to feed the learning algorithm with labeled instances where it is least confident about its own predictions. This \emph{uncertainty sampling} algorithm \cite{LewisGale:94} is a popular choice for active learning for NLP tasks with expensive gold labeling, such as text classification \cite{LewisGale:94}, word-sense disambiguation \cite{ChenETAL:06} and statistical parsing \cite{TangETAL:02}.
Our method falls into the category of stream-based online active learning (as opposed to pool-based active learning, selecting instances from a large pool of unlabeled data), since the algorithm decides on the fly (online) which translation prefixes of the stream of source tokens to request feedback for. Instead of receiving gold annotations, as in the studies mentioned above, our algorithm receives weaker, bandit feedback --- but the motivation of minimizing human labeling effort is the same.

%% file: sec-interactive.tex
In the following, we will introduce the key ideas of our approach, formalize them, and present an algorithm for reinforcement learning for interactive-predictive NMT.

\subsection{Actor-Critic Reinforcement Learning for NMT}

The objective of reinforcement learning methods is to maximize the expected reward obtainable from interactions of an agent (here: a machine translation system) with an environment (here: a human translator). In our case, the agent/system  performs actions by predicting target words $y_t$ according to a stochastic policy $p_\theta$ parameterized by an RNN encoder-decoder NMT system \cite{BahdanauETAL:15} where
\begin{align}
\label{eq:nmt}
p_{\theta}(\mathbf{y} | \mathbf{x}) = \prod^{T_y}_{t=1} p_{\theta}(y_t | \mathbf{x}, \mathbf{y}_{<t}).
\end{align}
The environment/human can be formalized as a Markov Decision Process where a state at time $t$ is a tuple $s_t = \left< \mathbf{x}, \mathbf{y}_{<t}\right>$ consisting of the conditioning context of the input $\mathbf{x}$ and the current produced history of target tokens $\mathbf{y}_{<t}$. Note that since states $s_{t+1}$ include the current chosen action $y_t$ and can contain long histories $\mathbf{y}_{<t}$, the state distribution is sparse and deterministic. The reward distribution of the environment/critic is estimated by function approximation in actor-critic methods. The reward estimator (called critic) is trained on actual rewards and updated after every interaction, and then used to update the parameters of the policy (called actor) in a direction of function improvement. We use the advantage actor critic framework of Mnih et al. \shortcite{MnihETAL:16} which estimates the advantage $A_{\phi}(y_t | s_t)$ in reward of choosing action $y_t$ in a given state $s_t$ over the mean reward value for that state. This framework has been applied to reinforcement learning for NMT by Nguyen et al. \shortcite{NguyenETAL:17}. The main objective of the actor is then to maximize the expected advantage
\begin{align}
\label{eq:actor}
L_\theta = \E_{p(\mathbf{x})p_{\theta}({\mathbf{y}} | \mathbf{x})} \left[ \sum_{t=1}^{T_y} A_{\phi}(y_t | s_t) \right].
\end{align}
The stochastic gradient of this objective for a sampled target word $\hat{y}_t$ for an input $\mathbf{x}$ can be calculated following the policy gradient theorem \cite{SuttonETAL:00,KondaTsitsiklis:00} as
\begin{align}
\label{eq:sgd}
\nabla L_\theta(\hat{y}_t) = \sum_{t=1}^{T_y} \left[ \nabla \log  p_{\theta}(\hat{y}_t | s_t) A_{\phi}(\hat{y}_t | s_t) \right].
\end{align}
In standard actor-critic algorithms, the parameters of actor and the critic are updated online at each time step. The actor parameters $\theta$ are updated by sampling $\hat{y}_t$ from $p_{\theta}$ and performing a step in the opposite direction of the stochastic gradient of $L_\theta(\hat{y}_t)$; the critic parameters $\phi$ are updated by minimizing $L_\phi(\hat{y}_t)$, defined as the mean squared error of the reward estimator for sampled target word $\hat{y}_t$ with respect to actual rewards (for more details see Nguyen et al. \shortcite{NguyenETAL:17}). In our experiments, we simulate user rewards by character F-score (chrF) values of partial translations.

\subsection{Triggering Human Feedback Requests by Actor Entropy}

Besides the idea of replacing human post-edits by human rewards, another key feature of our approach is to minimize the number of requests for human feedback. This is achieved by computing the uncertainty of the policy distribution as the average word-level entropy $\bar{H}$ of an $n$-word partial translation, defined as
\begin{equation}
\label{eq:H_ph}
\bar{H}(\hat{y}_{1:n}) = \frac{1}{n}\sum\limits_{t=1}^{n}\left[-\sum\limits_{v \in \mathcal{V}}p_{\theta}(v | s_{t}) \log p_{\theta}(v | s_{t}) \right],
\end{equation}
where $\hat{y}_{1:n} = \{\hat{y}_{1}, \hat{y}_{2}, \ldots, \hat{y}_{n}\}$ is a sequence of $n$ predicted tokens starting at the sentence beginning, $\mathcal{V}$ is the output vocabulary, and $p_{\theta}(v | s_{t})$ is the probability of predicting a word in $\mathcal{V}$ at state $s_t$ of the RNN decoder.

A request for human feedback is triggered when $\bar{H}(\hat{y}_{1:n})$ is higher than a running average $\gamma$ by a factor of $\epsilon$ or when $<$eos$>$ is generated. Upon receiving a reward from the user, both actor and critic are updated. Hence, our algorithm takes the middle ground between updating at each time step $t$ and performing an update only after a reward signal for the completed translation is received. In our simulation experiments, this process is repeated until the $<$eos$>$ token is generated, or when a pre-defined maximum length, here $T_{\max}=50$, is reached.

\subsection{Simulating Human Rewards on Translation Quality}

Previous work on reinforcement learning in machine translation has simulated human bandit feedback by evaluating full-sentence translations against references using per-sentence approximations of BLEU \cite{SokolovETALnips:16,KreutzerETAL:17,NguyenETAL:17}. We found that when working with partial translations, user feedback on translation quality can successfully be simulated by computing the chrF-score \cite{Popovic:15} of the translation with respect to the reference translation truncated to the same length. If the length of the translation exceeds the length of the reference, no truncation is used. We denote rewards as a function $\textrm{R}(\hat{y}_{1:t})$ of only the partial translation $\hat{y}_{1:t}$, in order to highlight the fact that rewards are in principle independent of reference translations.

\subsection{Sampling versus Forced Decoding via {Prefix Buffer $\Xi$}}

The standard approach to estimate the expected reward in policy gradient techniques is to employ Monte-Carlo methods, in specific, multinomial sampling of actions. This guarantees an unbiased estimator and allows sufficient exploration of the action space in learning. In contrast, interactive-predictive machine translation usually avoids exploration in favor of exploitation by decoding the best partial translation under the current model after every interaction. Since in our framework, learning and decoding are interleaved, we have to find the best compromise between exploration and exploitation.

The general modus operandi of our framework is simultaneous exploration and exploitation by multinomial sampling actions from the current policy. However, in cases where a partial translation receives a high user reward, we store it in a so-called {prefix buffer} $\Xi$, and perform forced decoding by feeding the prefix to the decoder for the remaining translation process.

\begin{algorithm}[t]
\caption{Bandit Interactive-Predictive}
\label{alg:BIP-NMT}
\begin{algorithmic}[1]
\STATE \textbf{Input:} $\theta_{0}$, $\phi_{0}$, $\alpha_A$, $\alpha_C$, $\epsilon$
\STATE \textbf{Output:} Estimates $\theta^\ast$, $\phi^\ast$
\STATE $k \leftarrow 1$
\FOR{i $\leftarrow$ 1, $\ldots$ N}
	\STATE Receive $\mathbf{x}_{i}$, Initialize $\gamma_{0} \leftarrow 0$, $\Xi \leftarrow \emptyset$
	\FOR{t $\leftarrow$ 1 $\ldots$ $T_{\max}$}
		\STATE Sample $\hat{y}_{1:t} \sim p_{\theta_{k-1}}(\cdot | \mathbf{x}_{i},\mathbf{\hat{y}}_{<t}, \Xi)$
		\STATE Compute $\bar{H}(\hat{y}_{1:t})$ using Eq. \eqref{eq:H_ph}
		\IF{$\bar{H}(\hat{y}_{1:t}) - \gamma_{t-1} \geq \epsilon \cdot \gamma_{t-1}$ \\
		 or $<$eos$>$ in $\hat{y}_{1:t}$}
			\STATE Receive feedback $\textrm{R}(\hat{y}_{1:t})$
				\IF{$\textrm{R}(\hat{y}_{1:t})\geq \mu$}
					\STATE $\Xi \leftarrow \hat{y}_{1:t}$
				\ENDIF
			    \STATE $\theta_k \leftarrow \theta_{k-1} -\alpha_A \nabla L_{\theta_{k-1}}(\hat{y}_{1:t})$ \\
			    (Eq. \eqref{eq:sgd})
				\STATE $\phi_k \leftarrow \phi_{k-1} -\alpha_C \nabla L_{\phi_{k-1}}(\hat{y}_{1:t})$ \\
				(see Eq. (7) in Nguyen et al. \shortcite{NguyenETAL:17})
				\STATE $k \leftarrow k + 1$
		\ENDIF
		\STATE Update $\gamma_{t} \leftarrow \gamma_{t-1}  + \frac{1}{t} \left( \bar{H}(\hat{y}_{1:t}) - \gamma_{t-1} \right) $
		\STATE \textbf{break} if $<$eos$>$ in $\hat{y}_{1:t}$
	\ENDFOR
\ENDFOR
\end{algorithmic}
\end{algorithm}

\subsection{Algorithm for Bandit Interactive-Predictive NMT}

Algorithm \ref{alg:BIP-NMT} gives pseudo-code for \textbf{B}andit-\textbf{I}nteractive-\textbf{P}redictive \textbf{N}eural \textbf{M}achine \textbf{T}ranslation (BIP-NMT). The algorithm receives an input source sequence $\mathbf{x_{i}}$ (line 4), and incrementally predicts a sequence of output target tokens up to length $T_{\max}$ (line 6). At each step $t$, a partial translation $\hat{y}_{1:t}$ is sampled from the policy distribution $p_{\theta}(\cdot|\mathbf{x_{i}},\mathbf{y}_{<t}, \Xi)$ that implements an RNN encoder-decoder with an additional prefix buffer $\Xi$ for forced decoding (line 7). User feedback is requested in case the average entropy $\bar{H}(\hat{y}_{1:t})$ of the policy is larger than or equal to a running average by a factor of $\epsilon$ or when $<$eos$>$ is generated (line 9). If the reward $\textrm{R}(\hat{y}_{1:t})$ is larger than or equal to a threshold $\mu$, the prefix is stored in a buffer for forced decoding (lines 11-12). Next, updates of the parameters of the policy (line 14), critic (line 15), and average entropy (line 17) are performed. Actor and critic each use a separate learning rate schedule ($\alpha_A$ and $\alpha_C$). 

\begin{figure}[t]
\center
	\includegraphics[width=0.8\columnwidth]{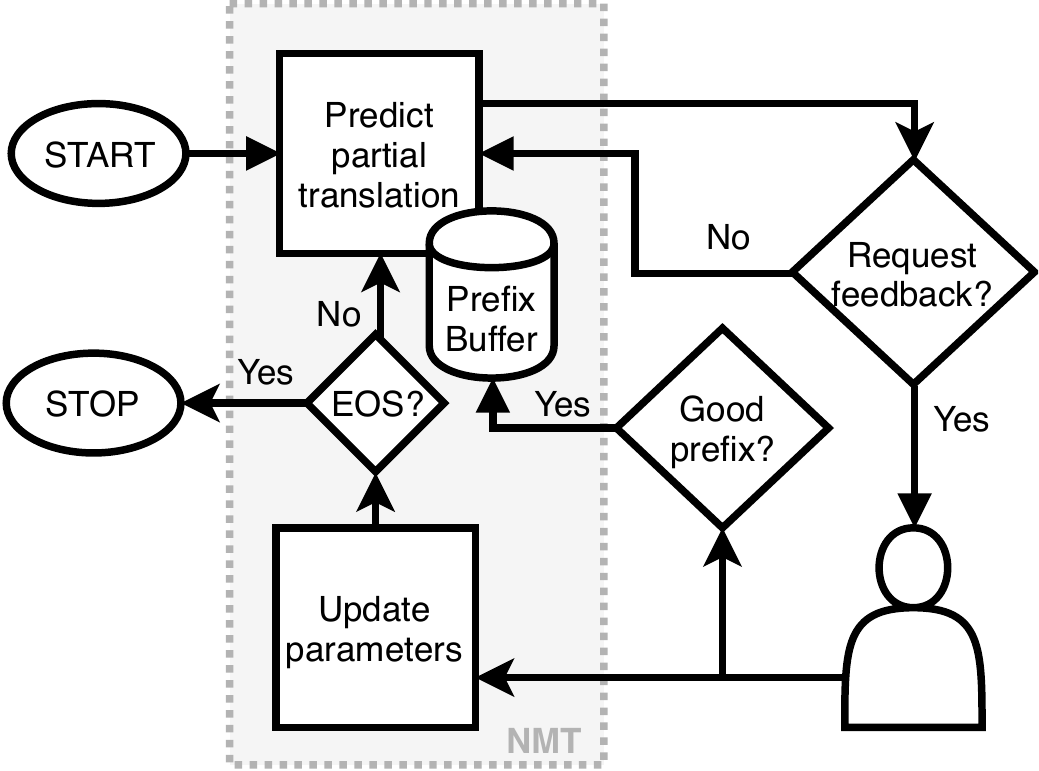}
	\caption{Interaction of the NMT system with the human during learning for a single translation.}
	\label{fig:interaction}
\end{figure}

Figure \ref{fig:interaction} visualizes the interaction of the BIP-NMT system with a human for a single translation: Feedback is requested when the model is uncertain or the translation is completed. It is directly used for a model update and, in case it was good, for filling the prefix buffer, before the model moves to generating the next (longer) partial translation.

%% file: sec-exps.tex
We simulate a scenario where the learning NMT system requests online bandit feedback for partial translations from a human in the loop. The following experiments will give an initial practical assessment of our proposed interactive learning algorithm. Our analysis of the interactions between actor, critic and simulated human will provide further insights into the learning behavior of BIP-NMT.

\subsection{Setup}

\paragraph{Data and Preprocessing.} We conduct experiments on French-to-English translation on Europarl (EP) and News Commentary (NC) domains. The large EP parallel corpus is used to pre-train the actor in a fully-supervised setting with a standard maximum likelihood estimation objective. The critic network is not pre-trained. For interactive training with bandit feedback, we extract 10k sentences from the NC corpus. Validation and test sets are also chosen from the NC domain. Note that in principle more sentences could be used, however, we would like to simulate a realistic scenario where human feedback is costly to obtain.
Data sets were tokenized and cleaned using Moses tools \cite{KoehnETAL:07}. Furthermore,  sentences longer than 50 tokens were removed from the training data. Each language's vocabulary contains the 50K most frequent tokens extracted from the two training sets. Table \ref{tab:data} summarizes the data statistics.
\begin{table}
	\centering
        \resizebox{\columnwidth}{!}{
	\begin{tabular}{lcccc}
		\toprule
		\textbf{Dataset} & \textbf{EP (v.5)} & $\bar{n}$ & \textbf{NC (WMT07)} & $\bar{n}$ \\
		\midrule
		Training (filt.) & 1,346,679 & 23.5 & 9,216 & 21.9 \\
		Validation 	& 2,000 & 29.4 & 1,064 & 24.1\\
		Test & - & -	& 2,007 & 24.8 \\
		\bottomrule
	\end{tabular}
        }
\caption{Number of parallel sentences and average number of words per sentence in target language (en), denoted by $\bar{n}$, for training (filtered to a maximum length of 50), validation and test sets for French-to-English translation for Europarl (EP) and News Commentary (NC) domains.}
\label{tab:data}
\end{table}

\paragraph{Model Configuration and Training.}
Following Nguyen et al. \shortcite{NguyenETAL:17}, we employ an architecture of two independent but similar encoder-decoder frameworks for actor and critic, respectively, each using global-attention \cite{LuongETAL:15} and unidirectional single-layer LSTMs\footnote{Our code can be accessed via the link \url{https://github.com/heidelkin/BIPNMT}.}. Both the size of word embedding and LSTM's hidden cells are 500. We used the Adam Optimizer \cite{KingmaBa:15} with $\beta_{1}= 0.9 $ and $\beta_{2}= 0.999$. During supervised pre-training, we train with mini-batches of size 64, and set Adam's $\alpha=10^{-3}$. A decay factor of 0.5 is applied to $\alpha$, starting from the fifth pass, when perplexity on the validation set increases. During interactive training with bandit feedback, we perform true online updates (i.e. mini-batch size is 1) with Adam's $\alpha$ hyper-parameter kept constant at $10^{-5}$ for both the actor and the critic. In addition, we clip the Euclidean norm of gradients to 5 in all training cases.

\begin{table*}[t]
	\centering
	\begin{tabular}{lcccc}
		\toprule
		\textbf{System} & \textbf{chrF (std)} & \textbf{BLEU (std)} & $\Delta$ \textbf{chrF} & $\Delta$ \textbf{BLEU} \\
		\midrule
		Out-of-domain NMT & 61.30 & 24.77 & 0 & 0\\
		Nguyen et al. \shortcite{NguyenETAL:17} & 62.25 (0.08) & 25.32 (0.02) & +0.95 & +0.55\\
		\textbf{BIP-NMT} ($\epsilon$ = 0.75, $\mu$ = 0.8) & 63.34 (0.12) & 26.95 (0.12) & +2.04 & +2.18 \\
		\bottomrule
	\end{tabular}
\caption{Evaluation of pre-trained out-of-domain baseline model, actor-critic learning on one epoch of sentence-level in-domain bandit feedback \cite{NguyenETAL:17} and BIP-NMT with settings $\epsilon = 0.75$, $\mu = 0.8$ trained on one epoch of sub-sentence level in-domain bandit feedback. Results are given on the NC test set according to average sentence-level chrF and corpus-level BLEU. Result differences between all pairs of systems are statistically significant according to \texttt{multeval} \cite{ClarkETAL:11}.}
\label{tab:test}
\end{table*}

\paragraph{Baselines and Evaluation.}
Our supervised out-of-domain baseline consists of the actor NMT system described as above, pre-trained on Europarl, with optimal hyperparameters chosen according to corpus-level BLEU on the validation set.
Starting from this pre-trained EP-domain model, we further train a bandit learning baseline by employing Nguyen's \shortcite{NguyenETAL:17} actor-critic model, trained on one epoch of sentence-level simulated feedback. The choice of comparing models after one epoch of training is a realistic simulation of a human-system interaction on a sequence of data where each input is seen only once. The feedback signal is simulated with chrF, using character-n-grams of length 6 and a value of $\beta=2$ of the importance factor of recall over precision. While during training exploration through sampling is essential, during inference and for final model evaluation we use greedy decoding. We evaluate the trained models on our test set from the NC-domain using average sentence-level chrF and standard corpus-level BLEU \cite{Papineni:02} to measure how well they got adapted to the new domain.

\begin{figure}[t]
	\includegraphics[width=1\linewidth]{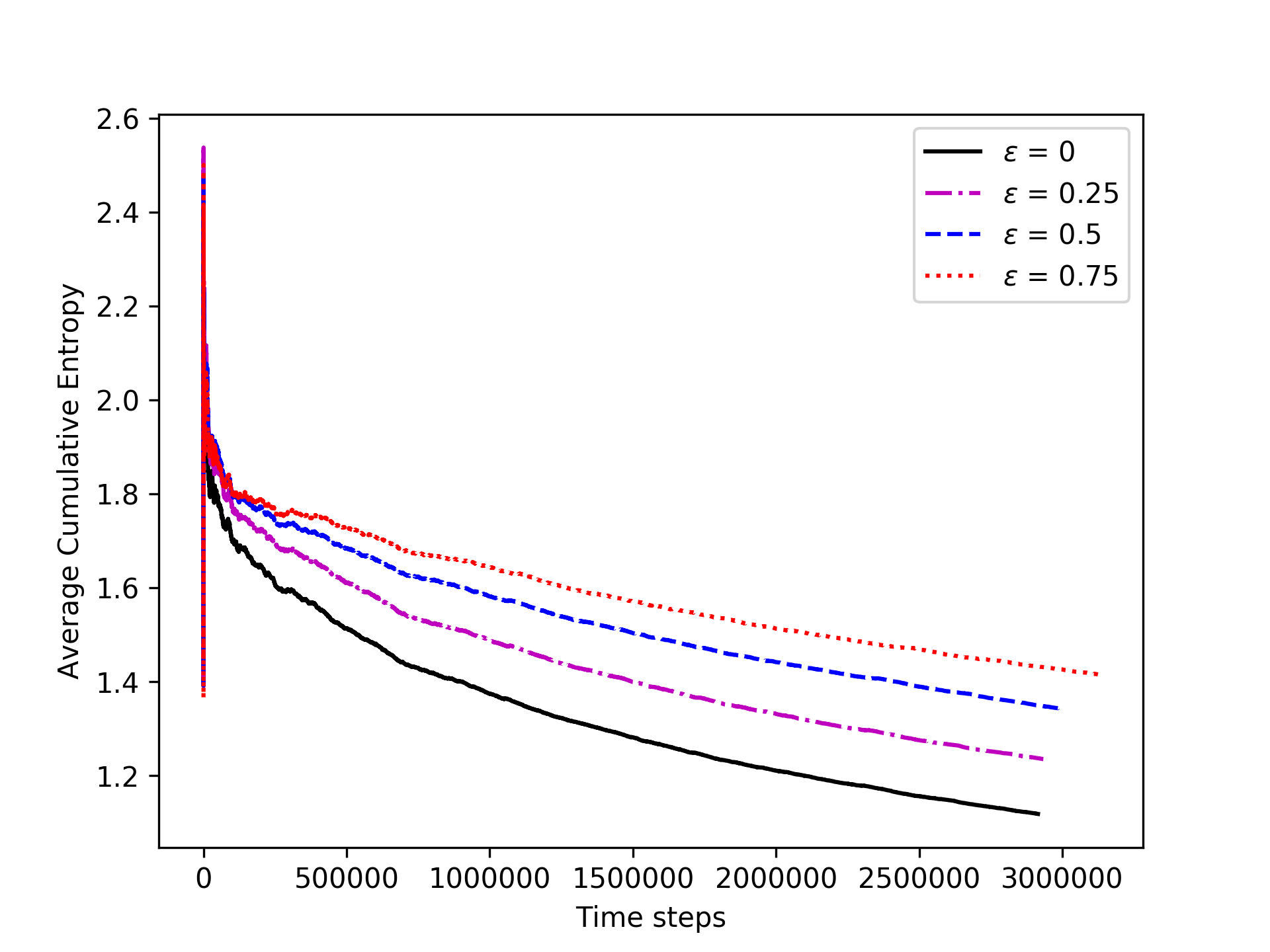}
	\caption{Average cumulative entropy during one epoch of BIP-NMT training with $\mu=0.8$ and $\epsilon = \{0, 0.25, 0.5, 0.75\}$.}
	\label{fig:entropy-curve}
\end{figure}

\subsection{Results and Analysis}

Table \ref{tab:test} shows the results of an evaluation of a baseline NMT model pre-trained by maximum likelihood on out-of-domain data. This is compared to an actor-critic baseline that trains the model of Nguyen et al. \shortcite{NguyenETAL:17} on sentence-level in-domain bandit feedback for one epoch. This approach can already improve chrF (+0.95) and BLEU (+0.55) significantly by seeing bandit feedback on in-domain data. BIP-NMT, with optimal hyperparameters $\epsilon = 0.75$, $\mu = 0.8$ chosen on the validation set, is trained in a similar way for one epoch, however, with the difference that even weaker sub-sentence level bandit feedback is provided on average 5 times per input. We see that BIP-NMT significantly improves both BLEU (+2.18) and chrF (+2.04) by even larger margins.

\begin{table*}[t]
	\centering
	\begin{tabular}{lllcccc}
		\toprule
		$\epsilon$ & \textbf{chrF (std)} & \textbf{BLEU (std)} &  \textbf{Avg \# Requests} & $\Delta$ \textbf{chrF} & $\Delta$ \textbf{BLEU} & $\Delta$ \textbf{Avg \# Requests} \\
		\midrule
		0   & 61.86 (0.06) & 25.54 (0.17) & 15.91 (0.01) & 0 & 0 & 0\\
		  
			0.25	 & 62.15 (0.17) & 25.84 (0.13) & 11.06 (0.07) & +0.29 & +0.3 & -5 \\
			
			0.5	 & 61.95 (0.05) & 25.46 (0.09) & 7.26 (0.03) & +0.09 & -0.08 & -9 \\
			
			0.75& 62.15 (0.04) & 25.07 (0.12) & 4.94 (0.02) & +0.29 & -0.47 & -11 \\
		\bottomrule
	\end{tabular}
	\caption{Impact of entropy margin $\epsilon$ on average sentence-level chrF score, corpus BLEU and average number of feedback requests per sentence on the NC validation set. The feedback quality threshold $\mu$ is set to 0.8 for all models.}
\label{tab:BIP-NMT-Valid}
\end{table*}

Table \ref{tab:BIP-NMT-Valid} analyzes the impact of the metaparameter $\epsilon$ of the BIP-NMT algorithm. We run each experiment three times and report mean results and standard deviation. $\epsilon$ controls the margin by which the average word-level entropy needs to increase with respect to the running average in order to trigger a feedback request. Increasing this margin from $0$ to $0.25$, $0.5$ and $0.75$ corresponds to decreasing the number of feedback requests by a factor of $3$ from around $16$ to around $5$. This reduction corresponds to a small increase in chrF (+0.29) and a small decrease in BLEU (-0.47). 

Figure \ref{fig:entropy-curve} shows another effect of the metaparameter $\epsilon$: It shows the variation of the average word-level entropy $\bar{H}$ over time steps of the algorithm during one epoch of training. This is computed as a cumulative average, i.e., the value of $\bar{H}$ is accumulated and averaged over the number of target tokens produced for all inputs seen so far. We see that average cumulative entropy increases in the beginning of the training, but then decreases rapidly, with faster rates for smaller values of $\epsilon$, corresponding to more updates per input.

The metaparameter $\mu$ controls the threshold of the reward value that triggers a reuse of the prefix for forced decoding.  In our experiments, we set this parameter to a value of $0.8$ in order to avoid re-translations of already validated prefixes, even if they might sometimes lead to better final full translations. We found the effect of lowering $\mu$ from $1.0$ to $0.8$ negligible on the number of feedback requests and on translation quality but beneficial for the usability.

\begin{table*}[th]
\resizebox{\textwidth}{!}{%

\begin{tabular}{ll}
\toprule

\multicolumn{2}{l}{\textbf{SRC} \makecell[cl]{depuis 2003 , la chine est devenue le plus important partenaire commercial du mexique apr\`{e}s les etats-unis .}}\\
\multicolumn{2}{l}{\textbf{REF} \makecell[cl]{since 2003 , china has become mexico 's most important trading partner after the united states . $</$s$>$} }\\
\\
\textbf{Partial sampled translation} & \textbf{Feedback} \\
since & 1\\
\underline{since} 2003 , china has & 1\\
\underline{since 2003 , china has} become & 1\\
\underline{since 2003 , china has become} mexico & 1\\
\underline{since 2003 , china has become mexico} 's & 1\\
\underline{since 2003 , china has become mexico 's} most & 1\\
\underline{since 2003 , china has become mexico 's most} important & 1\\
\makecell[cl]{\underline{since 2003 , china has become mexico 's most important} trading partner\\ after the us . $</$s$>$ }& 0.8823
\\
\midrule

\\
\midrule
\multicolumn{2}{l}{\textbf{SRC} \makecell[cl]{la r\'{e}ponse que nous , en tant qu' individus , acceptons est que nous sommes libres parce que nous nous gouvernons\\ nous-m\^{e}mes en commun plut\^{o}t que d' \^{e}tre dirig\'{e}s par une organisation qui n' a nul besoin de tenir compte de notre existence .}}\\
\multicolumn{2}{l}{\textbf{REF} \makecell[cl]{the answer that we as individuals accept is that we are free because we rule ourselves in common , \\rather than being ruled by some agency that need not take account of us . $</$s$>$}} \\
\\
\textbf{Partial sampled translation} & \textbf{Feedback} \\
the & 1\\
\underline{the} answer & 1\\
\underline{the answer} we & 0.6964\\
\underline{the answer} we , & 0.6246 \\
\underline{the answer} \textcolor{red}{we} as individuals allow to 14 are & 0.6008\\
\makecell[cl]{\underline{the answer}  \textcolor{red}{we , as individuals , go} down \textcolor{red}{to speak 8 , are being} free \textcolor{red}{because we} govern ourselves\\ \textcolor{red}{, rather from being} based \textcolor{red}{together}} & 0.5155\\
 \makecell[cl]{\underline{the answer} \textcolor{red}{we} , as people , accepts is that we principle are free because we govern ourselves ,\\ \textcolor{red}{rather than} being led by a organisation which has absolutely no need to take our standards . $</$s$>$} & 0.5722\\
\midrule
\\

\\
\midrule
\multicolumn{2}{l}{\textbf{SRC} \makecell[cl]{lors d' un rallye ``journ\'{e}e j\'{e}rusalem'' tenu \`{a} l' universit\'{e} de t\'{e}h\'{e}ran en d\'{e}cembre 2001 , il a prononc\'{e} l' une des menaces\\ les plus sinistres du r\'{e}gime .}}\\
\multicolumn{2}{l}{\textbf{REF} \makecell[cl]{at a jerusalem day rally at tehran university in december 2001 , he uttered one of the regime 's most sinister threats . $</$s$>$}}\\
\\
\textbf{Partial sampled translation} & \textbf{Feedback} \\
\textcolor{red}{in} & 0 \\
 \makecell[cl]{\textcolor{red}{in a} round of jerusalem called a academic university in teheran in december 2001 \textcolor{red}{,}\\ he declared one in the most recent hostility \textcolor{red}{to} the regime \textcolor{red}{ . $</$s$>$}} & 0.5903 \\
\bottomrule
\end{tabular}%
}
\caption{Interaction protocol for three translations. These translations were sampled from the model when the algorithm decided to request human feedback (line 10 in Algorithm \ref{alg:BIP-NMT}). Tokens that get an overall negative reward (in combination with the critic), are marked in red, the remaining tokens receive a positive reward. When a prefix is good (i.e. $\geq \mu$, here $\mu=0.8$) it is stored in the buffer and used for forced decoding for later samples (underlined).}
\label{tab:example}
\end{table*}

\subsection{Example Protocols} Table \ref{tab:example} presents user-interaction protocols for three examples encountered during training of BIP-NMT with $\epsilon=0.75, \mu=0.8$. For illustrative purposes, we chose examples that differ with respect to the number of feedback requests, the use of the prefix buffer, and the feedback values.  Prefixes that receive a feedback $\geq \mu$ and are thus stored in the buffer and re-used for later samples are indicated by underlines. Advantage scores $<0$ indicate a discouragement of individual tokens and are highlighted in red.

In the first example, the model makes frequent feedback requests (in 8 of 17 decoding steps) and fills the prefix buffer due to the high quality of the samples.  The second example can use the prefix buffer only for the first two tokens since the feedback varies quite a bit for subsequent partial translations. Note how the token-based critic encourages a few phrases of the translations, but discourages others. The final example shows a translation where the model is very certain and hence requests feedback only after the first and last token (minimum number of feedback requests). The critic correctly identifies problematic parts of the translation regarding the choice of prepositions.

%% file: sec-disc.tex
We presented a novel algorithm, coined BIP-NMT, for bandit interactive-predictive NMT using reinforcement learning techniques. Our algorithm builds on advantage actor-critic learning \cite{MnihETAL:16,NguyenETAL:17} for an interactive translation process with a human in the loop. The advantage over previously presented algorithms for interactive-predictive NMT is the low human effort for producing feedback (a translation quality judgment instead of a correction of a translatioin), even further reduced by an active learning strategy to request feedback only for situations where the actor is uncertain.

We showcased the success of BIP-NMT with simulated feedback, with the aim of moving to real human feedback in future work. Before deploying this algorithm in the wild, suitable interfaces for giving real-valued feedback have to be explored to create a pleasant user experience. Furthermore, in order to increase the level of human control, a combination with the standard paradigm that allows user edits might be considered in future work.

Finally, our algorithm is in principle not limited to the application of NMT, but can furthermore --- thanks to the broad adoption of neural sequence-to-sequence learning in NLP --- be extended to other structured prediction or sequence generation tasks.

%% file: paper.bbl
\begin{thebibliography}{}

\bibitem[\protect\citename{Bahdanau \bgroup et al.\egroup
  }2015]{BahdanauETAL:15}
Bahdanau, Dzmitry, Kyunghyun Cho, and Yoshua Bengio.
\newblock 2015.
\newblock Neural machine translation by jointly learning to align and
  translate.
\newblock In {\em Proceedings of the International Conference on Learning
  Representations {(ICLR)}}, San Diego, {CA}.

\bibitem[\protect\citename{Bahdanau \bgroup et al.\egroup
  }2017]{BahdanauETAL:17}
Bahdanau, Dzmitry, Philemon Brakel, Kelvin Xu, Anirudh Goyal, Ryan Lowe, Joelle
  Pineau, Aaron Courville, and Yoshua Bengio.
\newblock 2017.
\newblock An actor-critic algorithm for sequence prediction.
\newblock In {\em Proceedings of the 5th International Conference on Learning
  Representations {(ICLR)}}, Toulon, France.

\bibitem[\protect\citename{Barrachina \bgroup et al.\egroup
  }2008]{BarrachinaETAL:08}
Barrachina, Sergio, Oliver Bender, Francisco Casacuberta, Jorge Civera, Elsa
  Cubel, Shahram Khadivi, Antonio Lagarda, Hermann Ney, Jes{\'u}s Tom{\'a}s,
  Enrique Vidal, and Juan-Miguel Vilar.
\newblock 2008.
\newblock Statistical approaches to computer-assisted translation.
\newblock {\em Computational Linguistics}, 35(1):3--28.

\bibitem[\protect\citename{Bentivogli \bgroup et al.\egroup
  }2016]{BentivogliETAL:16}
Bentivogli, Luisa, Nicola Bertoldi, Mauro Cettolo, Marcello Federico, Matteo
  Negri, and Marco Turchi.
\newblock 2016.
\newblock On the evaluation of adaptive machine translation for human
  post-editing.
\newblock {\em {IEEE} Transactions on Audio, Speech and Language Processing
  {(TASLP))}}, 24(2):388--399.

\bibitem[\protect\citename{Chen \bgroup et al.\egroup }2006]{ChenETAL:06}
Chen, Jinying, Andrew Schein, Lyle Ungar, and Martha Palmer.
\newblock 2006.
\newblock An empirical study of the behavior of active learning for word sense
  disambiguation.
\newblock In {\em Human Language Technologies: The 2006 Annual Conference of
  the North American Chapter of the Association for Computational Linguistics
  {(NAACL-HLT)}}, New York City, NY.

\bibitem[\protect\citename{Clark \bgroup et al.\egroup }2011]{ClarkETAL:11}
Clark, Jonathan, Chris Dyer, Alon Lavie, and Noah Smith.
\newblock 2011.
\newblock Better hypothesis testing for statistical machine translation:
  Controlling for optimizer instability.
\newblock In {\em Proceedings of the 49th Annual Meeting of the Association for
  Computational Linguistics {(ACL'11)}}, Portland, {OR}.

\bibitem[\protect\citename{Domingo \bgroup et al.\egroup }2018]{DomingoETAL:18}
Domingo, Miguel, {\'A}lvaro Peris, and Francisco Casacuberta.
\newblock 2018.
\newblock Segment-based interactive-predictive machine translation.
\newblock {\em Machine Translation}.

\bibitem[\protect\citename{Foster \bgroup et al.\egroup }1997]{FosterETAL:97}
Foster, George, Pierre Isabelle, and Pierre Plamondon.
\newblock 1997.
\newblock Target-text mediated interactive machine translation.
\newblock {\em Machine Translation}, 12(1-2):175--194.

\bibitem[\protect\citename{Foster \bgroup et al.\egroup }2002]{FosterETAL:02}
Foster, George, Philippe Langlais, and Guy Lapalme.
\newblock 2002.
\newblock User-friendly text prediction for translators.
\newblock In {\em Proceedings of the Conference on Empirical Methods in Natural
  Language Processing {(EMNLP)}}, Philadelphia, {PA}.

\bibitem[\protect\citename{Gonz{\'a}lez-Rubio \bgroup et al.\egroup
  }2011]{Gonzalez-RubioETAL:11}
Gonz{\'a}lez-Rubio, Jes{\'u}s, Daniel Ortiz-Mart{\'\i}nez, and Francisco
  Casacuberta.
\newblock 2011.
\newblock An active learning scenario for interactive machine translation.
\newblock In {\em Proceedings of the 13th International Conference on
  Multimodal Interfaces {(ICMI)}}, Barcelona, Spain.

\bibitem[\protect\citename{Gonz{\'a}lez-Rubio \bgroup et al.\egroup
  }2012]{Gonzalez-RubioETAL:12}
Gonz{\'a}lez-Rubio, Jes{\'u}s, Daniel Ortiz-Mart{\'\i}nez, and Francisco
  Casacuberta.
\newblock 2012.
\newblock Active learning for interactive machine translation.
\newblock In {\em Proceedings of the 13th Conference of the European Chapter of
  the Association for Computational Linguistics {(EACL)}}, Avignon, France.

\bibitem[\protect\citename{Green \bgroup et al.\egroup }2014]{GreenETAL:14}
Green, Spence, Sida~I. Wang, Jason Chuang, Jeffrey Heer, Sebastian Schuster,
  and Christopher~D. Manning.
\newblock 2014.
\newblock Human effort and machine learnability in computer aided translation.
\newblock In {\em Proceedings the Conference on Empirical Methods in Natural
  Language Processing {(EMNLP)}}, Doha, Qatar.

\bibitem[\protect\citename{Judah \bgroup et al.\egroup }2010]{JudahETAL:10}
Judah, Kshitij, Saikat Roy, Alan Fern, and Thomas~G. Dietterich.
\newblock 2010.
\newblock Reinforcement learning via practice and critique advice.
\newblock In {\em Proceedings of the 24th {AAAI} Conference on Artificial
  Intelligence}, Atlanta, {GA}.

\bibitem[\protect\citename{Karimova \bgroup et al.\egroup
  }2017]{KarimovaETAL:17}
Karimova, Sariya, Patrick Simianer, and Stefan Riezler.
\newblock 2017.
\newblock A user-study on online adaptation of neural machine translation to
  human post-edits.
\newblock {\em CoRR}, abs/1712.04853.

\bibitem[\protect\citename{Kingma and Ba}2015]{KingmaBa:15}
Kingma, Diederik~P. and Jimmy Ba.
\newblock 2015.
\newblock Adam: A method for stochastic optimization.
\newblock In {\em Proceedings of the International Conference on Learning
  Representations {(ICLR)}}, San Diego, {CA}.

\bibitem[\protect\citename{Knowles and Koehn}2016]{KnowlesKoehn:16}
Knowles, Rebecca and Philipp Koehn.
\newblock 2016.
\newblock Neural interactive translation prediction.
\newblock In {\em Proceedings of the Conference of the Association for Machine
  Translation in the Americas {(AMTA)}}, Austin, {TX}.

\bibitem[\protect\citename{Koehn \bgroup et al.\egroup }2007]{KoehnETAL:07}
Koehn, Philipp, Hieu Hoang, Alexandra Birch, Chris Callison-Birch, Marcello
  Federico, Nicola Bertoldi, Brooke Cowan, Wade Shen, Christine Moran, Richard
  Zens, Chris Dyer, Ondrej Bojar, Alexandra Constantin, and Evan Herbst.
\newblock 2007.
\newblock Moses: Open source toolkit for statistical machine translation.
\newblock In {\em Proceedings of the {ACL} 2007 Demo and Poster Sessions},
  Prague, Czech Republic.

\bibitem[\protect\citename{Konda and Tsitsiklis}2000]{KondaTsitsiklis:00}
Konda, Vijay~R. and John~N. Tsitsiklis.
\newblock 2000.
\newblock Actor-critic algorithms.
\newblock In {\em Advances in Neural Information Processing Systems {(NIPS)}},
  Vancouver, Canada.

\bibitem[\protect\citename{Kreutzer \bgroup et al.\egroup
  }2017]{KreutzerETAL:17}
Kreutzer, Julia, Artem Sokolov, and Stefan Riezler.
\newblock 2017.
\newblock Bandit structured prediction for neural sequence-to-sequence
  learning.
\newblock In {\em Proceedings of the 55th Annual Meeting of the Association for
  Computational Linguistics {(ACL)}}, Vancouver, Canada.

\bibitem[\protect\citename{Lewis and Gale}1994]{LewisGale:94}
Lewis, David~D and William~A Gale.
\newblock 1994.
\newblock A sequential algorithm for training text classifiers.
\newblock In {\em Proceedings of the 17th Annual International ACM-SIGIR
  Conference on Research and Development in Information Retrieval {(SIGIR)}},
  Dublin, Ireland.

\bibitem[\protect\citename{Luong \bgroup et al.\egroup }2015]{LuongETAL:15}
Luong, Thang, Hieu Pham, and Christopher~D. Manning.
\newblock 2015.
\newblock Effective approaches to attention-based neural machine translation.
\newblock In {\em Proceedings of the Conference on Empirical Methods in Natural
  Language Processing {(EMNLP)}}, Lisbon, Portugal.

\bibitem[\protect\citename{Marie and Max}2015]{MarieMax:15}
Marie, Benjamin and Aur{\'e}lien Max.
\newblock 2015.
\newblock Touch-based pre-post-editing of machine translation output.
\newblock In {\em Proceedings of the Conference on Empirical Methods in Natural
  Language Processing {(EMNLP)}}, Lisbon, Portugal.

\bibitem[\protect\citename{Mnih \bgroup et al.\egroup }2016]{MnihETAL:16}
Mnih, Volodymyr, Adri{\`{a}}~Puigdom{\`{e}}nech Badia, Mehdi Mirza, Alex
  Graves, Timothy~P. Lillicrap, Tim Harley, David Silver, and Koray
  Kavukcuoglu.
\newblock 2016.
\newblock Asynchronous methods for deep reinforcement learning.
\newblock In {\em Proceedings of the 33rd International Conference on Machine
  Learning {(ICML)}}, New York, {NY}.

\bibitem[\protect\citename{Nguyen \bgroup et al.\egroup }2017]{NguyenETAL:17}
Nguyen, Khanh, Hal Daum\'{e}, and Jordan Boyd-Graber.
\newblock 2017.
\newblock Reinforcement learning for bandit neural machine translation with
  simulated feedback.
\newblock In {\em Proceedings of the Conference on Empirical Methods in Natural
  Language Processing {(EMNLP)}}, Copenhagen, Denmark.

\bibitem[\protect\citename{Ortiz-Mart{\'i}nez \bgroup et al.\egroup
  }2010]{Ortiz-MartinezETAL:10}
Ortiz-Mart{\'i}nez, Daniel, Ismael Garc{\'i}a-Varea, and Francisco Casacuberta.
\newblock 2010.
\newblock Online learning for interactive statistical machine translation.
\newblock In {\em Human Language Technologies: The 2010 Annual Conference of
  the North American Chapter of the Association for Computational Linguistics
  {(NAACL-HLT)}}, Los Angeles, {CA}.

\bibitem[\protect\citename{Papineni \bgroup et al.\egroup }2002]{Papineni:02}
Papineni, Kishore, Salim Roukos, Todd Ward, and Wei-Jing Zhu.
\newblock 2002.
\newblock Bleu: a method for automatic evaluation of machine translation.
\newblock In {\em Proceedings of the 40th Annual Meeting on Association for
  Computational Linguistics {(ACL)}}, Philadelphia, {PA}.

\bibitem[\protect\citename{Popovi{\'c}}2015]{Popovic:15}
Popovi{\'c}, Maja.
\newblock 2015.
\newblock chr{F}: character n-gram f-score for automatic mt evaluation.
\newblock In {\em Proceedings of the Tenth Workshop on Statistical Machine
  Translat ion {(WMT)}}, Lisbon, Portugal.

\bibitem[\protect\citename{Sokolov \bgroup et al.\egroup
  }2016]{SokolovETALnips:16}
Sokolov, Artem, Julia Kreutzer, Christopher Lo, and Stefan Riezler.
\newblock 2016.
\newblock Stochastic structured prediction under bandit feedback.
\newblock In {\em Advances in Neural Information Processing Systems {(NIPS)}},
  Barcelona, Spain.

\bibitem[\protect\citename{Sutton and Barto}2017]{SuttonBarto:17}
Sutton, Richard~S. and Andrew~G. Barto.
\newblock 2017.
\newblock {\em Reinforcement Learning. An Introduction}.
\newblock The {MIT} Press, second edition.

\bibitem[\protect\citename{Sutton \bgroup et al.\egroup }2000]{SuttonETAL:00}
Sutton, Richard~S., David McAllester, Satinder Singh, and Yishay Mansour.
\newblock 2000.
\newblock Policy gradient methods for reinforcement learning with function
  approximation.
\newblock In {\em Advances in Neural Information Processings Systems {(NIPS)}},
  Vancouver, Canada.

\bibitem[\protect\citename{Tang \bgroup et al.\egroup }2002]{TangETAL:02}
Tang, Min, Xiaoqiang Luo, and Salim Roukos.
\newblock 2002.
\newblock Active learning for statistical natural language parsing.
\newblock In {\em Proceedings of the 40th Annual Meeting of the Association for
  Computational Linguistics (ACL)}, Pennsylvania, PA.

\bibitem[\protect\citename{Williams}1992]{Williams:92}
Williams, Ronald~J.
\newblock 1992.
\newblock Simple statistical gradient-following algorithms for connectionist
  reinforcement learning.
\newblock {\em Machine Learning}, 8:229--256.

\bibitem[\protect\citename{Wuebker \bgroup et al.\egroup }2016]{WuebkerETAL:16}
Wuebker, Joern, Spence Green, John DeNero, Sasa Hasan, and Minh-Thang Luong.
\newblock 2016.
\newblock Models and inference for prefix-constrained machine translation.
\newblock In {\em Proceedings of the 54th Annual Meeting of the Association for
  Computational Linguistics {(ACL)}}, Berlin, Germany.

\end{thebibliography}
